\documentclass{article}

\usepackage{arxiv}
\usepackage[utf8]{inputenc} 
\usepackage[T1]{fontenc}    
\usepackage{xcolor}
\usepackage{hyperref}       
\usepackage{url}            
\usepackage{booktabs}       
\usepackage{amsfonts}       
\usepackage{amsmath}
\usepackage{microtype}      
\usepackage{graphicx}
\usepackage[numbers,sort&compress]{natbib}
\usepackage{doi}
\usepackage{tabularx}
\usepackage{xspace}
\usepackage{float}
\usepackage{algorithm}
\usepackage{listings}
\usepackage{algpseudocode}

\lstdefinestyle{jsoncompact}{
  basicstyle=\ttfamily\scriptsize,
  columns=fullflexible,
  keepspaces=true,
  breaklines=true,
  showstringspaces=false
}

\hypersetup{
  colorlinks=true,
  linkcolor=blue!50!black,
  citecolor=blue!50!black,
  urlcolor=blue!50!black
}

\newcolumntype{Y}{>{\raggedright\arraybackslash}X}

\newcommand{\tactile}{\textsc{Tactile}\xspace}
\newcommand{\ax}{\textsc{AX}\xspace}
\newcommand{\ocr}{\textsc{OCR}\xspace}
\newcommand{\mcp}{\textsc{MCP}\xspace}
\newcommand{\sys}{\tactile}

\title{\tactile: Giving Computer-Using Agents Hands and Feet}
\author{
  Yong Liu$^{1}$, Zhenyi Zhong$^{1}$, Zhanpeng Shi$^{1}$\\
  $^{1}$Shanghai Jiao Tong University\\
  liuyong9975@sjtu.edu.cn \\
  \url{https://github.com/yliust/Tactile}
}
\date{}

\renewcommand{\headeright}{}
\renewcommand{\undertitle}{}
\renewcommand{\shorttitle}{\tactile}

\hypersetup{
  pdftitle={Tactile: Giving Computer-Using Agents Hands and Feet},
  pdfauthor={Yong Liu, Zhenyi Zhong, Zhanpeng Shi},
  pdfkeywords={computer-use agents, accessibility, GUI automation, OCR, MCP, open source},
}

\begin{document}
\maketitle

\begin{abstract}
Computer-use agents are becoming capable software operators, but their interface to desktop applications is still often a brittle motor layer: they look at screenshots, predict coordinates, click, and hope that the visible state changed as intended. This collapses target grounding, action execution, and outcome verification into a single ambiguous operation. We present \sys, an open-source tool layer that gives agents a more reliable ``hands and feet'' for desktop use. \sys converts heterogeneous UI evidence--operating-system accessibility semantics, \ocr-grounded text, and visual fallback regions--into action-grounded interface states: compact target candidates with source labels, roles or text, state, geometry, executable affordances, and verification cues. Agents operate through an observe-ground-act-verify loop that prefers native semantic actions when available, falls back to \ocr-grounded coordinates when visible text is the best evidence, and keeps full provenance for replay and failure attribution. On macOSWorld-style tasks, adding \sys improves Codex Success@100 from 41.1\% to 50.0\% overall and from 45.2\% to 55.3\% on accessibility-adapted tasks; a 96-task cross-agent subset shows consistent gains across Codex, Claude Code, OpenCode, and Goose. These results suggest that reliable computer use requires not only stronger models, but also a reusable execution substrate that exposes software actions as semantic, verifiable, and auditable objects rather than anonymous screen coordinates.
\begin{figure}[h]
  \centering
  \includegraphics[width=0.5\columnwidth]{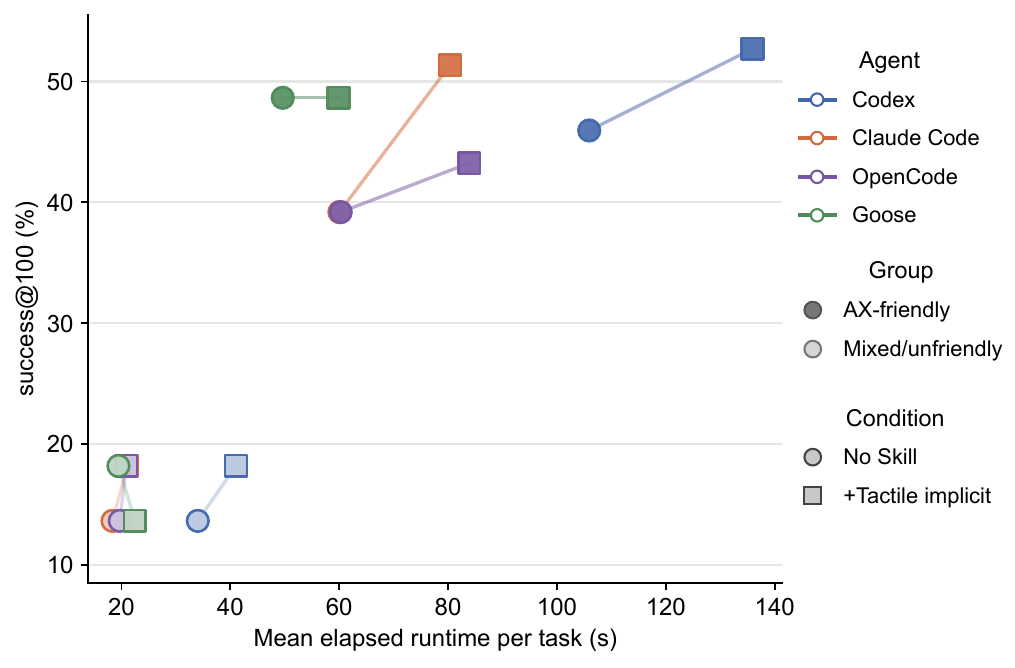}
  \caption{Success rate versus average task runtime for four agents, with and without \tactile, across accessibility-friendly and accessibility-limited tasks. }
  \label{fig:teaser}
\end{figure}
\end{abstract}

\section{Introduction}

Large language model agents are moving from text-only assistance into live software environments. Web agents operate dynamic websites, mobile agents interact with applications, and desktop agents attempt open-ended tasks in full operating systems~\cite{deng2023mind2web,zheng2024seeact,he2024webvoyager,niu2024screenagent,xie2024osworld,rawles2024androidworld,macosworld2025}. In these environments, intelligence alone is not enough. A useful agent also needs an operating body: a way to perceive what can be touched, choose a target, execute the right primitive, and verify that the software moved into the intended state.

The dominant body for many computer-use agents is screenshot-first control. The agent observes a rendered screen, uses visual reasoning or \ocr to localize relevant text and regions, predicts low-level mouse or keyboard actions, and then observes the next screen state~\cite{zheng2024seeact,he2024webvoyager,seeclick2024,niu2024screenagent,omniparser2024}. This interface is general and important: it works even when an application exposes no useful programmatic structure. However, it also hides the executable structure of software. A screenshot can show the word ``Send,'' but it usually does not say whether the word belongs to an enabled button, whether the button supports a native press action, which control will receive focus, or what state change should be used to verify success.

\begin{figure*}[h]
  \centering
  \includegraphics[width=0.98\textwidth]{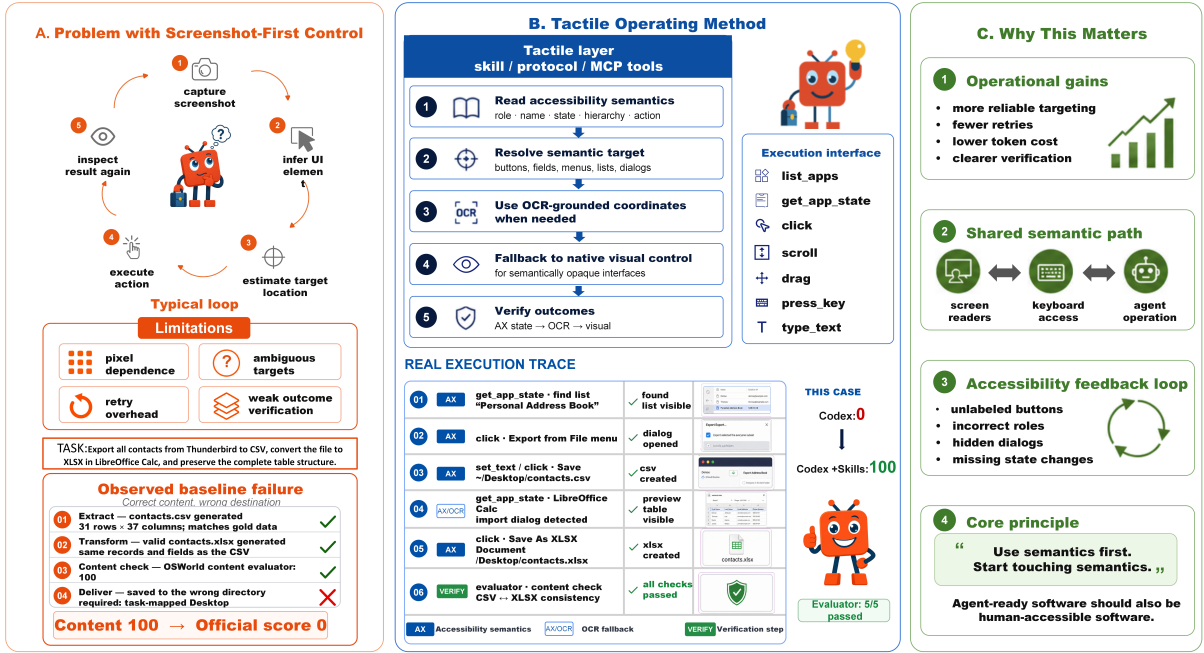}
  \caption{Motivation and overview. Screenshot-first control is general but often entangles perception, target selection, low-level motor action, and verification. \tactile inserts an action-grounded operating layer: it reads accessibility semantics when available, uses \ocr-grounded text when necessary, falls back to visual control for semantically opaque regions, and records enough provenance to make each operation inspectable.}
  \label{fig:teaser}
\end{figure*}

This paper starts from a simple observation: many graphical applications already expose structured interface semantics for screen readers, keyboard navigation, switch control, and other assistive technologies. On macOS, the Accessibility API represents UI elements through roles, names, values, hierarchy, frames, states, and actions~\cite{appleAccessibilityProgrammingGuide,appleAXHierarchy,appleAXUIElement,appleAXPerformAction}. Windows UI Automation exposes analogous properties and control patterns~\cite{microsoftUIAutomation,microsoftUIAControlPatterns}. Web accessibility standards define semantic requirements for perceivable and operable interfaces~\cite{w3cWCAG22,w3cWAIARIA12}. This infrastructure was designed for human accessibility, but it is also a useful execution substrate for agents.

Accessibility semantics should not be treated as a complete replacement for vision. Real applications often expose incomplete, noisy, stale, or misleading accessibility trees~\cite{screen2ax2025}. Custom canvases, remote desktops, media timelines, image-only controls, and webviews may require \ocr or visual grounding. The design question is therefore not whether agents should use accessibility \emph{or} vision. The question is how an agent runtime should order and combine evidence so that each step uses the richest available signal: stable semantic targets when software exposes them, current \ocr coordinates when visible text is the best locator, and visual fallback when the interface is semantically opaque.

We present \sys, an open-source \mcp-compatible tool layer~\cite{mcpSpec,mcpToolsSpec} for accessibility-aware desktop operation. \sys is not a new foundation model. Instead, it provides a reusable operating interface that existing agent clients can call. The central abstraction is an \emph{action-grounded interface state}: a compact set of target candidates that preserve what a UI object is, where it is, what actions it supports, which evidence produced it, and how the next observation may verify an outcome. The full observation is retained outside the model-facing context for replay, debugging, and failure attribution.

Our contributions are:
\begin{enumerate}
  \item We formulate computer use as an action-grounded interface problem in which targetability, actionability, verifiability, and auditability are first-class runtime properties.
  \item We introduce an accessibility-first operating ladder that composes accessibility semantics, \ocr-grounded coordinates, and visual fallback without reducing the agent to an \ax-only system.
  \item We implement \sys as an open \mcp tool layer with \texttt{get\_app\_state}, semantic and coordinate action tools, normalized coordinate contracts, compact candidate generation, and trace provenance.
  \item We evaluate \sys on macOSWorld-style tasks and show preliminary gains for Codex and for a 96-task cross-agent subset covering Codex, Claude Code, OpenCode, and Goose.
\end{enumerate}

\section{Related Work}
\label{sec:background}

\subsection{Computer Use as Grounded Interaction}

Agentic systems commonly interleave reasoning and acting~\cite{yao2022react}. For computer-use agents, acting requires grounding natural-language intent in a graphical environment whose state is large, dynamic, partially observable, and controlled through low-level actions. Prior work has explored web pages, mobile devices, and full operating-system environments~\cite{deng2023mind2web,zheng2024seeact,he2024webvoyager,niu2024screenagent,xie2024osworld,rawles2024androidworld,macosworld2025}.

Rendered observations provide a strong default substrate. They are visually faithful, framework-agnostic, and compatible with applications that expose little programmatic structure. Screenshot-based agents often enrich raw images with \ocr boxes, visual element proposals, and coordinate systems~\cite{seeclick2024,omniparser2024,niu2024screenagent}. Native GUI-agent models such as UI-TARS further emphasize end-to-end screen perception, grounding, action modeling, and reasoning over GUI trajectories~\cite{qin2025uitars}. These approaches are essential because many UI regions cannot be represented reliably by structured APIs.

The limitation is that rendered observations do not directly expose execution semantics. They usually do not encode whether a control is enabled, selected, expanded, focused, or invokable through a native action. They also conflate three different decisions: which UI object is intended, which motor primitive should be executed, and which post-action evidence confirms success. This makes failures hard to attribute. A failed step may come from stale observation, ambiguous target selection, coordinate conversion, focus mismatch, application non-response, or weak verification.

\subsection{Accessibility Semantics as Agent Infrastructure}

Accessibility APIs decompose graphical interfaces into semantic elements~\cite{appleAccessibilityProgrammingGuide,appleAXHierarchy,appleAXUIElement,microsoftUIAutomation,w3cWAIARIA12}. A typical accessible element may expose a role, such as button or text field; a readable name, value, description, or placeholder; state, such as enabled, focused, selected, expanded, or settable; hierarchy and path information; supported actions; and screen geometry.

These fields are directly useful for agents because they separate \emph{what} an element is from \emph{where} it is rendered. When a button supports \texttt{AXPress}, the runtime can invoke a native semantic action instead of synthesizing a click at a transient screen point~\cite{appleAXUIElement,appleAXPerformAction}. After acting, the agent can re-observe whether a checkbox is selected, a text field value has changed, a dialog has closed, or a new list item has appeared. Accessibility semantics are not complete enough to be the only interface, but they are useful enough to be the first interface when available.

\subsection{Tool Layers, Skills, and Reusable Computer-Use Substrates}

A complementary line of work studies how agents should access software through tools, skills, or structured runtime interfaces. MCP-style tool protocols standardize how agents call external capabilities~\cite{mcpSpec,mcpToolsSpec}. Skill-centric systems such as CUA-Skill encode reusable human computer-use procedures as parameterized skills and execution graphs~\cite{chen2026cuaskill}. These systems point to the same broader need: agents should not repeatedly rediscover low-level interaction procedures from raw pixels.

\sys occupies a different layer. It does not provide a large library of task-specific skills, and it does not replace model-side planning. Instead, it exposes a local execution substrate that makes each desktop state more actionable. A skill, planner, or language agent can call the same \mcp observation and action tools, receive compact target candidates, and execute through semantic actions or coordinate fallbacks under the same provenance model.

\subsection{Design Requirements}

The motivation above leads to four requirements for an agent-facing computer-use layer:
\begin{table}[H]
  \centering
  \small
  \begin{tabularx}{\linewidth}{@{}lY@{}}
    \toprule
    Requirement & Meaning for desktop operation \\
    \midrule
    Targetability & The observation should expose candidate UI objects or text regions that can be selected deliberately, not only pixels. \\
    Actionability & The runtime should expose safe executable affordances, such as native press actions, typing, scrolling, dragging, or coordinate clicks. \\
    Verifiability & Each state-changing action should be followed by evidence that can support or weaken the success claim. \\
    Auditability & The system should retain provenance so developers can inspect what was observed, selected, executed, and verified. \\
    \bottomrule
  \end{tabularx}
  \caption{Runtime requirements that motivate \tactile. The goal is to make computer use inspectable at the level of interface objects, actions, and outcomes rather than only screenshots and coordinates.}
  \label{tab:requirements}
\end{table}

\section{Methods: The \tactile Operating Model}
\label{sec:methods}

\subsection{Action-Grounded Interface States}

\tactile represents a desktop observation as an action-grounded interface state. The state is a ranked set of target candidates produced from multiple evidence sources. A candidate may be backed by an accessibility element, an \ocr line, a visual fallback region, or a merge of several sources. Each candidate carries a short identifier, source labels, role or recognized text, state, geometry, executable affordances, and provenance.

This representation changes the model-facing contract. The agent is not asked to choose an anonymous coordinate first. It can choose a target hypothesis such as ``the enabled Send button from accessibility, strengthened by an overlapping OCR label,'' while the runtime chooses the safest executable primitive. If the candidate supports a semantic press action, \tactile can invoke that action. If the target is only an \ocr line, the runtime can click the current text center. If no structured evidence is reliable, the agent can still fall back to visual control.

\subsection{The Accessibility-First Operating Ladder}

\tactile orders available evidence through a ladder rather than a mutually exclusive choice between accessibility and vision:
\begin{description}
  \item[Level 1: Accessibility semantics.] Use roles, names, values, states, hierarchy, frames, source handles, and native actions when applications expose them. This level gives the strongest action and verification contracts.
  \item[Level 2: \ocr-grounded coordinates.] Use current \ocr lines when visible text is the most reliable target signal or when an accessible element lacks a useful name. OCR evidence is coordinate-backed and source-labeled, not treated as semantic state.
  \item[Level 3: Visual fallback.] Use screenshot-based or model-native visual control for semantically opaque regions such as canvases, media editors, remote desktops, icon-only widgets, and misleading accessibility trees.
\end{description}

The ladder is accessibility-first, not accessibility-only. Its purpose is to prefer the richest available structure before resorting to less structured signals.

\subsection{Observe-Ground-Act-Verify}

The operating loop separates four decisions that screenshot-first control often collapses:
\begin{description}
  \item[Observe.] Collect accessibility elements, \ocr lines, optional visual metadata, active application information, coordinate contracts, and a compact model-facing summary.
  \item[Ground.] Select a candidate with explicit source labels, state, geometry, actions, and provenance. Candidate choice is separated from motor execution.
  \item[Act.] Prefer native semantic actions when safe and available. Otherwise use text input, keyboard input, scrolling, dragging, or coordinate actions from the latest observation.
  \item[Verify.] Re-observe after state-changing actions and check state changes, text changes, dialog transitions, list mutations, or visual evidence. When verification is weak, record that weakness instead of silently treating the step as confirmed.
\end{description}

This loop gives the agent a stable operating rhythm. A model can say which candidate it intends to use; the runtime can decide whether that candidate should be pressed, focused, typed into, scrolled, dragged, clicked, or rejected as stale; and the trace can explain what happened.

\section{System Design and Implementation}
\label{sec:system}

\begin{figure*}[t]
  \centering
  \includegraphics[width=0.98\textwidth]{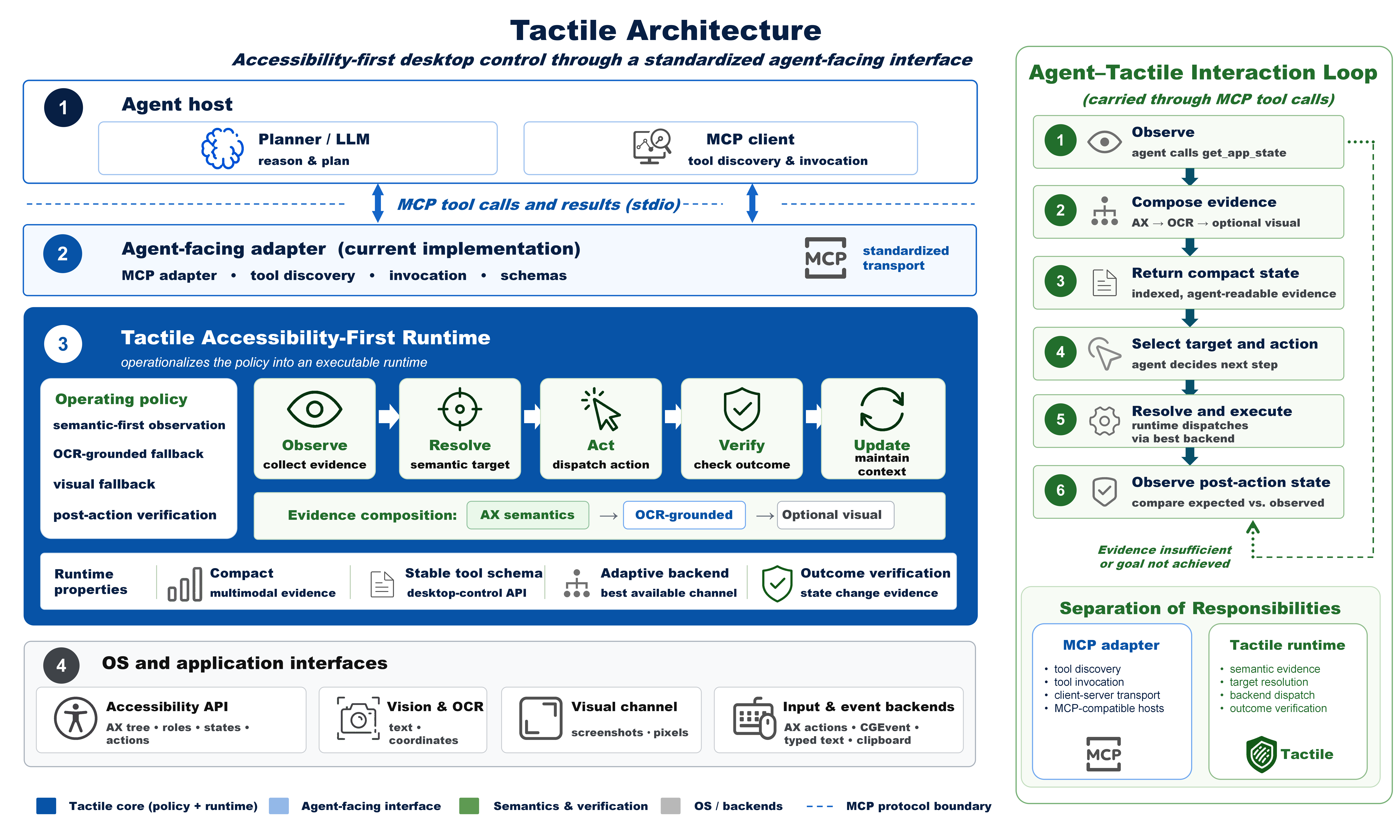}
  \caption{\tactile system architecture. Agent clients call a standardized \mcp tool surface. The \tactile server compiles raw accessibility, \ocr, and visual evidence into compact observations, executes actions through a local macOS runtime, and returns verification evidence while preserving full traces for replay and debugging.}
  \label{fig:architecture}
\end{figure*}

Figure~\ref{fig:architecture} shows the implementation structure. \tactile is exposed as an \mcp server~\cite{mcpSpec,mcpToolsSpec}. Agent clients interact with a stable tool surface rather than depending on platform-specific accessibility calls. The current implementation is strongest on macOS: a Swift runtime handles Accessibility traversal and actions, ScreenCapture, Vision \ocr, and input events~\cite{appleAXUIElement,appleAXPerformAction,appleScreenCaptureKit,appleVisionRecognizeText,appleQuartzEventServices}; a Python layer handles tool schemas, candidate compaction, workflow state, and trace logging. The repository also includes early Windows support following the same abstraction through UI Automation~\cite{microsoftUIAutomation,microsoftUIAControlPatterns}.

\subsection{From Skill Policy to Executable MCP Tools}

The tool layer exposes reusable primitives including \texttt{list\_apps}, \texttt{get\_app\_state}, \texttt{perform\_secondary\_action}, \texttt{click}, \texttt{type\_text}, \texttt{press\_key}, \texttt{scroll}, and \texttt{drag}. The policy encoded around these tools is deliberately conservative: resolve the active application, inspect structured accessibility evidence before escalating to visual evidence, prefer semantic actions when safe, re-observe after state-changing actions, and split high-risk external actions into locate, draft, verify, and submit phases.

This separation keeps planning and execution modular. A planner or skill can decide what should happen next, while \tactile provides a client-agnostic schema and runtime guarantees for how a candidate is represented and how an action is executed.

\subsection{The \texttt{get\_app\_state} Evidence Compiler}
\label{sec:get-app-state}

The central observation call is \texttt{get\_app\_state}. Rather than returning an unprocessed screenshot or a raw accessibility tree, the call acts as an evidence compiler. It transforms heterogeneous UI signals into a compact set of candidates useful for grounding, acting, and verification. Figure~\ref{fig:evidence-compiler} illustrates the pipeline.

\paragraph{Inputs.}
The compiler supports multiple observation modes. In \texttt{ax} mode, it traverses the accessibility tree and extracts semantic elements. In \texttt{ax\_ocr} mode, it additionally extracts \ocr lines with screen-space coordinates~\cite{appleVisionRecognizeText}. In \texttt{ax\_ocr\_visual} mode, it may attach visual context such as screenshot regions or visual fallback metadata. These modes let the caller trade context size against evidence richness.

\paragraph{Candidate construction.}
Accessibility elements contribute roles, names, values, states, hierarchy, frames, supported actions, and source handles. OCR lines contribute recognized text, bounding boxes, centers, and OCR provenance. Visual regions contribute image-space or screen-space regions when neither accessibility nor \ocr provides a reliable target. All coordinates are normalized into a common screen-point contract before candidates are exposed to the model.

\paragraph{Source-aware merging and ranking.}
The compiler merges evidence only when sources support a shared target hypothesis. An \ocr line overlapping an accessibility button may strengthen the candidate label, while an \ocr line inside a semantically opaque canvas remains a separate coordinate-backed candidate. Candidates are ranked by action-relevant properties such as visibility, enabled state, name quality, focus, action support, textual relevance, and recency. Ranking is a context-management mechanism, not a claim of guaranteed task success.

\paragraph{Output.}
The model-facing output contains a prioritized candidate list, global UI summary, active application metadata, coordinate contract, and verification cues. The runtime also writes a full-state dump for replay and failure analysis. The following example illustrates the candidate contract rather than a fixed implementation schema.

\begin{algorithm}[H]
\caption{\texttt{get\_app\_state}: Evidence Compiler for UI Target Construction}
\label{alg:get-app-state}
\footnotesize
\begin{algorithmic}[1]
\Require Active application $app$, observation mode $mode$
\Ensure Ranked target hypothesis set $H$ and compact state $S$
\State $E_A \gets \Call{CollectAccessibility}{app}$
\If{$mode = \texttt{ax}$}
  \State $E_O \gets \emptyset$; $E_V \gets \emptyset$
\Else
  \State $E_O \gets \Call{CollectOCR}{app}$
  \If{$mode = \texttt{ax\_ocr\_visual}$}
    \State $E_V \gets \Call{CollectVisual}{app}$
  \Else
    \State $E_V \gets \emptyset$
  \EndIf
\EndIf
\State $E \gets \Call{NormalizeGeometry}{E_A \cup E_O \cup E_V}$
\State $H \gets \Call{FormTargetHypotheses}{E}$
\State $H \gets \Call{ResolveCrossSourceIdentity}{H}$
\State $H \gets \Call{ScoreByActionabilityAndStability}{H}$
\State $S \gets \Call{EmitModelState}{H}$
\State \Call{WriteProvenanceLog}{$E, H, S$}
\State \Return $S$
\end{algorithmic}
\end{algorithm}

Algorithm~\ref{alg:get-app-state} summarizes the compiler as a dataflow contract. Platform adapters may differ in how they collect accessibility nodes, OCR lines, or visual regions, but the model-facing output must preserve the same invariants: source labels, normalized coordinates, action affordances, state, ranking, and provenance.

\subsection{Execution Runtime and Coordinate Contracts}

For actions, the runtime selects among semantic accessibility actions, CGEvent-based mouse and keyboard events, typed text, and clipboard-based fallbacks depending on the target and operation~\cite{appleAXUIElement,appleAXPerformAction,appleQuartzEventServices}. Public coordinates are macOS top-left screen points. \ocr lines include screen frames and centers. Visual screenshots report the screen region they cover. This normalized coordinate contract prevents a common failure mode in which agents mix screenshot pixels, retina-scaled pixels, and global screen coordinates.

\subsection{Context Optimization and Trace Instrumentation}

A practical desktop application may expose hundreds or thousands of accessibility nodes, many of which are decorative, redundant, off-screen, or irrelevant to the current task~\cite{screen2ax2025}. \sys therefore treats context management as a runtime design problem. The compact observation is not a lossy replacement for the full state; it is a model-facing view optimized for action selection. The full observation remains available for debugging, replay, and failure analysis.

Each run records observations, selected candidates, action plans, action results, verification summaries, confirmed facts, and failure categories. This trace structure makes it possible to inspect whether an intended target was absent from the observation, present but filtered out, retained but incorrectly ranked, selected correctly but executed through an unsuitable primitive, or executed successfully but not verified.

\section{Results}
\label{sec:results}

We evaluate \tactile on macOSWorld-style desktop tasks~\cite{macosworld2025} using two complementary views. The first view compares Codex~\cite{openaiCodexCLI} and Codex+\sys on the currently scored macOSWorld sample set. The second view uses a 96-task horizontal subset for cross-agent comparison across Codex, Claude Code, OpenCode, and Goose~\cite{openaiCodexCLI,anthropicClaudeCode,opencodeDocs,gooseDocs}. For each agent, we compare a no-skill baseline against a \emph{With Tactile} condition. The reported cross-agent \emph{With Tactile} score is a skill-optional upper bound computed from the best per-task result among no-skill, tactile-implicit, and tactile-explicit settings; it estimates the value of making \tactile available, not a learned routing policy.

\paragraph{Task splits and metric.}
We partition tasks into \emph{AX-adapted} tasks, where key interactive elements expose useful accessibility metadata, and \emph{Limited AX} tasks, where critical steps require \ocr or visual fallback because accessibility metadata is missing, incomplete, or unreliable. We report Success@100, the percentage of tasks completed within 100 interaction steps, following step-budgeted success-rate conventions in GUI-agent evaluation~\cite{xie2024osworld,macosworld2025}.

\subsection{Main Results}

\begin{figure*}[h]
  \centering
  \includegraphics[width=0.98\textwidth]{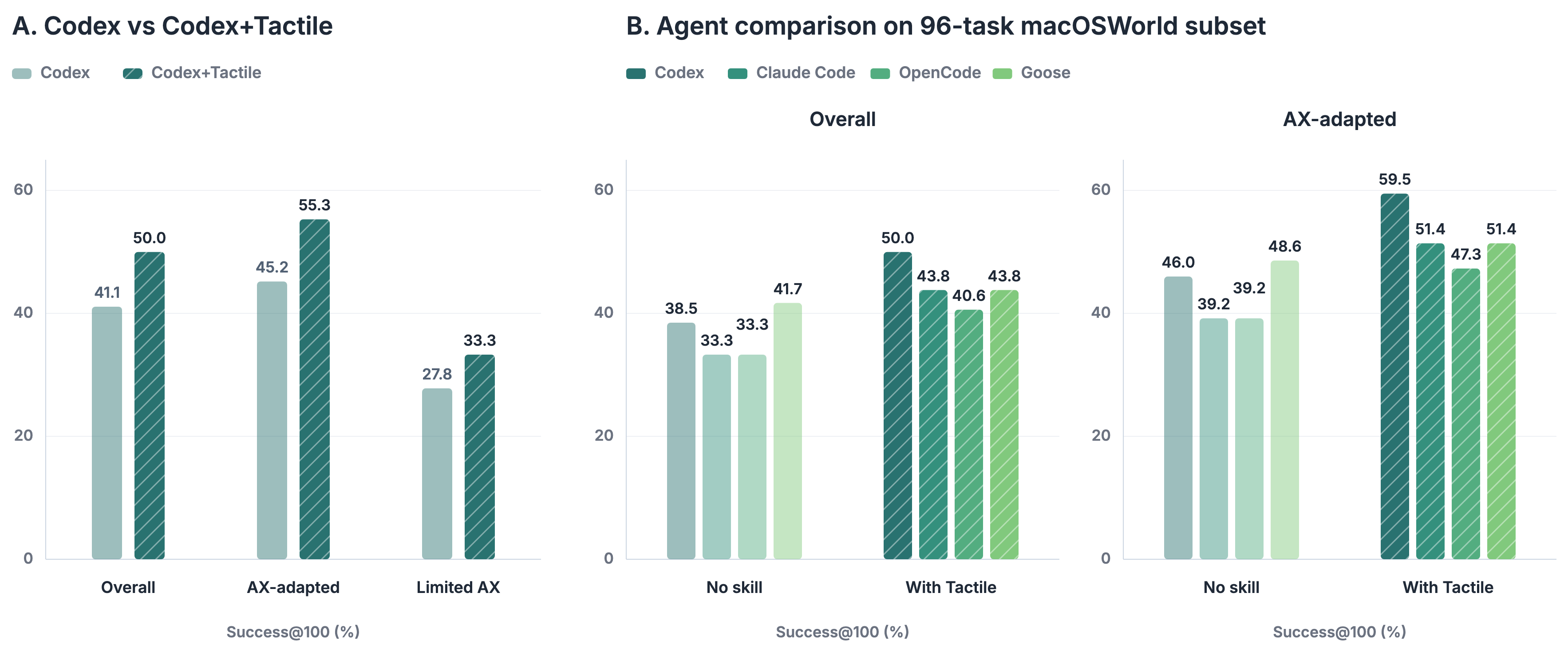}
  \caption{Benchmark results for \tactile on macOSWorld-style tasks. Left: Codex+\tactile improves Success@100 over Codex across the overall scored set, the accessibility-adapted split, and the limited-accessibility split. Right: on the 96-task horizontal subset, the overall and accessibility-adapted splits compare no-skill execution with the With Tactile condition for Codex, Claude Code, OpenCode, and Goose.}
  \label{fig:benchmark-results}
\end{figure*}

Figure~\ref{fig:benchmark-results} summarizes the benchmark results. In the graded Codex comparison, adding \tactile raises Success@100 from 41.06\% to 50.00\% overall, an 8.94 percentage-point gain. The gain is larger on AX-adapted tasks, rising from 45.22\% to 55.26\% (+10.04 points). On Limited-AX tasks, success improves from 27.78\% to 33.33\% (+5.55 points). This pattern is consistent with the operating ladder: accessibility semantics provide the strongest signal when available, while OCR and visual fallback remain useful for semantically weak interfaces.

The cross-agent subset shows the same qualitative pattern. Under the skill-optional upper bound, total success improves from 38.54\% to 50.00\% for Codex, 33.33\% to 43.75\% for Claude Code, 33.33\% to 40.62\% for OpenCode, and 41.67\% to 43.75\% for Goose. On AX-adapted tasks, Codex improves from 46.0\% to 59.5\%, Claude Code from 39.2\% to 51.4\%, OpenCode from 39.2\% to 47.3\%, and Goose from 48.6\% to 51.4\%. Gains are smaller on Limited-AX tasks, and Goose shows no change there, which reinforces the interpretation that \tactile is most valuable when applications expose actionable semantic structure.

\subsection{Trace-Based Examples}

The aggregate benchmark is complemented by trace inspection. In a recorded Zoom scheduling task, the user asked the agent to open Zoom, schedule a meeting for Friday night from 8pm to 9pm, set calendar options, and save the meeting. The trace contains eleven state snapshots. The initial application state retained 52 accessibility candidates after filtering. After the scheduling dialog opened, later snapshots contained hundreds of elements, including 412 elements after the schedule action and 468 elements with the date picker open. Traversal times in the snapshots ranged from 0.24s to 0.85s.

The important property is not that every element was accessible. Rather, the trace records which candidates were visible to the agent, which candidate was selected, which primitive was executed, which coordinate contract was used when applicable, and which evidence was available for verification. Table~\ref{tab:application-examples} summarizes additional applications used to characterize the operating ladder alongside benchmarked tasks.

\begin{table}[H]
  \centering
  \small
  \begin{tabularx}{\linewidth}{@{}lYY@{}}
    \toprule
    Application & Useful evidence & Fallback pressure and trace value \\
    \midrule
    Zoom & Dialogs, buttons, text fields, date picker elements, enabled states, and native actions. & Large trees require filtering and ranking; traces expose selected targets, dialog transitions, and post-action verification. \\
    Feishu/Lark & Navigation sections, search controls, organization buttons, compose fields, and application entries. & Mixed webview regions may require OCR; traces show whether workflow controls were selected by role, label, and state. \\
    WeChat & Some standard controls and navigation regions expose useful semantics. & Chat timelines, profile cards, and media-heavy regions often require OCR or visual fallback, which is explicit in the trace. \\
    CapCut/Jianying & Menus, dialogs, import/export controls, text fields, and some buttons. & Timelines, previews, trim handles, and media canvases require visual operation; traces mark the boundary between semantic and visual control. \\
    \bottomrule
  \end{tabularx}
  \caption{Trace-based examples illustrating different positions on the \tactile operating ladder.}
  \label{tab:application-examples}
\end{table}

\subsection{Benchmark Scope}

The benchmark is a system-level validation of the interface and operating ladder, not a complete claim about all desktop automation. It uses currently graded macOSWorld samples~\cite{macosworld2025} and a 96-task horizontal subset. The skill-optional cross-agent score is an upper bound rather than a learned routing policy. Future evaluations should expand the task set, report uncertainty estimates, compare against stronger visual-only baselines, and evaluate policies that decide when to invoke \tactile without oracle selection.

\section{Discussion}
\label{sec:discussion}

\subsection{What the Gains Mean}

The results support \tactile as an optional operating layer rather than an \ax-only replacement for visual computer use. Its main value is to expose richer target structure, executable affordances, coordinate contracts, and verification cues when software provides them. Screenshot and \ocr grounding remain essential because many applications have incomplete semantic layers or contain inherently visual regions~\cite{seeclick2024,omniparser2024,appleVisionRecognizeText,screen2ax2025}.

This framing avoids a false dichotomy. The question is not whether agents should use accessibility or vision. The question is which evidence should be used first for a given operation, how sources should be merged, how the resulting action should be executed, and how the outcome should be verified.

\subsection{Open-Source Tooling as Infrastructure}

\tactile is intended primarily as reusable infrastructure. When desktop-control stacks are tightly coupled to individual agent runtimes, results become difficult to reproduce and failures become difficult to compare. An open \mcp-compatible tool layer makes the interface explicit: different agent clients can call the same observation and action primitives, benchmark them on the same tasks, and improve platform adapters without changing the high-level policy.

\subsection{Agent-Ready Software Should Be Human-Accessible Software}

\tactile reframes accessibility metadata as shared infrastructure. For human users, accessibility semantics enable screen readers, keyboard navigation, switch control, and other assistive technologies~\cite{appleAccessibilityProgrammingGuide,w3cWCAG22,w3cWAIARIA12}. For agents, the same semantics provide structured observations, stable target identities, native action affordances, and state-based verification. If a button lacks a readable name, a screen-reader user suffers and an agent also struggles. If a dialog is invisible to the accessibility tree, both assistive technology and semantic agents lose access. This alignment creates a positive incentive: improving human accessibility also improves agent operability.

\section{Limitations and Future Work}
\label{sec:limitations}

\paragraph{Incomplete or noisy accessibility metadata.}
Accessibility metadata can be missing, incorrect, stale, or too noisy for direct use~\cite{screen2ax2025}. Some controls expose roles without names, some custom views expose large containers without useful children, and some actions are listed but do not produce reliable state changes. Future versions should add stronger consistency checks between accessibility state, OCR text, and visual evidence.

\paragraph{Cross-platform variation.}
Accessibility APIs differ across operating systems, application frameworks, sandboxing models, browser/webview implementations, and permission systems~\cite{appleAccessibilityProgrammingGuide,appleAXUIElement,microsoftUIAutomation,microsoftUIAControlPatterns,w3cWAIARIA12}. The current implementation is strongest on macOS. A complete cross-platform version will require platform-specific adapters that preserve the same candidate schema, coordinate contract, and provenance model.

\paragraph{Large and dynamic UI trees.}
Applications may expose hundreds or thousands of accessibility nodes, including decorative, redundant, off-screen, or stale elements~\cite{screen2ax2025}. Candidate compaction reduces model-facing context, but it can omit low-ranked elements that later become relevant. Future work should support query-driven expansion, hierarchical reveal, and task-conditioned re-ranking.

\paragraph{OCR and visual fallback failures.}
OCR fails on low-contrast text, icon-only controls, animations, overlapping windows, unusual fonts, and unsupported languages~\cite{long2019rethinking,appleVisionRecognizeText}. Visual fallback is necessary for canvas-like, drag-heavy, remote, or media-editing interfaces, but it reintroduces coordinate-level ambiguity. Future work should make fallback decisions more explicit by recording why structured evidence was insufficient and which fallback signal was used.

\paragraph{Verification remains difficult.}
Some applications provide no reliable semantic, textual, or visual feedback after an action. In such cases, the agent may be unable to determine whether an operation succeeded, failed, or is still pending. \sys currently records verification levels and failure categories, but stronger verification will require application-specific expectations, temporal reasoning, and possibly user confirmation for high-risk actions.

\section{Conclusion}
\label{sec:conclusion}

\tactile gives computer-using agents a reusable operating body: an open \mcp-compatible layer that turns desktop interfaces into action-grounded, verifiable, and auditable states. It does not replace screenshots, \ocr, or visual reasoning; instead, it orders them with accessibility semantics so that agents can use the richest available evidence at each step. By exposing targets as objects with state, actions, geometry, and provenance, \tactile makes desktop operation less like blind coordinate clicking and more like controlled interaction with software. The preliminary macOSWorld-style results show that this infrastructure can improve grounded execution across agents, especially when applications expose useful accessibility semantics.

\section*{Acknowledgements}

\tactile builds on decades of work from accessibility communities, screen-reader developers, operating-system accessibility APIs, \ocr systems, UI automation projects, agent runtimes, and open-source contributors. We also thank the maintainers of \texttt{mediar-ai/mcp-server-macos-use}~\cite{mediarMcpServerMacosUse}, whose open-source implementation informed parts of our code. The project reflects practical experience operating real macOS applications whose accessibility quality varies widely.

\bibliographystyle{unsrtnat}
\bibliography{references}

\end{document}